\documentclass{article}

\usepackage[final]{colm2026_conference}

\usepackage{microtype}
\usepackage{hyperref}
\usepackage{url}
\usepackage{graphicx}
\usepackage{booktabs}
\usepackage{multirow}
\usepackage{amsmath}
\usepackage{subcaption}
\usepackage{enumitem}
\usepackage{float}

\definecolor{darkblue}{rgb}{0, 0, 0.5}
\hypersetup{colorlinks=true, citecolor=darkblue, linkcolor=darkblue, urlcolor=darkblue}

\title{Automatic or Controlled? Repetition Priming Reveals Divergent Processing in Base LLMs, Instruct LLMs, and Humans}

\author{Jinglei Ren \\
School of Medicine \\
Yale University \\
\texttt{jinglei.ren@yale.edu} \\
\And
Yuyue Wang \\
Department of Computer Science \\
University of California, Los Angeles \\
\texttt{yuyue@cs.ucla.edu}
}

\begin{document}
\maketitle

\begin{abstract}
Words recur constantly in natural language use, yet it remains unclear whether language models reactivate prior representations or re-evaluate repeated words afresh, and whether post-training changes this default behavior. In this paper, we apply repetition priming \citep{shiffrin1977controlled} to 15 models across 5 families (1.5B--14B parameters) in two tasks (semantic categorization and cloze completion), with matched human experiments using identical stimuli. Our findings reveal that: (1) Base models exhibit \emph{automatic processing}, showing immediate facilitation that is stable across lags, partially survives context removal, and directly correlates with attention to prior occurrences; (2) Instruct models exhibit \emph{controlled processing}, showing facilitation that decays with lag, collapses without expected context, and at larger scales reverses to interference; (3) The dissociation widens monotonically with scale within the Qwen~2.5 family (1.5B--14B), suggesting that post-training progressively suppresses automatic facilitation; (4) Humans show a hybrid profile that is lag-sensitive like instruct models but uniformly facilitative like base models, never reversing to interference. Our study demonstrates that the post-training pipeline is associated with a qualitative shift in how models process repeated information, provides mechanistic evidence for where the two modes diverge, and reveals that human cognition occupies a middle ground that neither model type captures.
\end{abstract}

\section{Introduction}

Words recur constantly in natural language use, including in multi-turn dialogues, overlapping retrieval passages, and repeated few-shot exemplars \citep{brown2020language, ouyang2022training}. When a language model encounters a word it has already processed in the current context, does it simply reactivate the prior representation, or does it re-evaluate the word afresh? We present evidence that the answer is associated with post-training, including supervised instruction tuning and, for some models, preference optimization. Base models default to fast reactivation: repeating a word facilitates processing regardless of intervening context. Instruct models engage in context-dependent re-evaluation: facilitation decays with lag and, at scale, reverses to interference. The difference is not merely quantitative but reflects two qualitatively distinct processing modes.

To characterize these mechanisms, we draw on the \emph{automatic vs.\ controlled processing} framework from cognitive psychology \citep{shiffrin1977controlled}. Automatic processing is fast, stimulus-driven, and robust to interference; controlled processing is slower, context-dependent, and sensitive to disruption. Repetition priming, a well-established paradigm from human psycholinguistics \citep{scarborough1977frequency, forster1984repetition, tenpenny1995}, offers a controlled way to probe these mechanisms: by varying the lag between repeated exposures, we can test whether facilitation is durable and stimulus-driven or transient and sensitive to contextual interference. Recent work has begun applying dual-process frameworks to LLM behavior, framing the shift from fast heuristic responses to deliberate reasoning as a System~1 to System~2 transition \citep{brady2025dual, li2025system, ziabari2025reasoning}, but primarily through prompting manipulations. We take a different approach: using repetition priming to show that the post-training pipeline is associated with a qualitative shift in the default processing mode.

We test 15 models across 5 families (1.5B--14B parameters) in two complementary experiments: \emph{semantic categorization} (Experiment~1), which probes how repeated exposure affects forced-choice classification, and \emph{cloze completion} (Experiment~2; \citealp{taylor1953, peelle2020completion}), which probes how repeated exposure aids open-ended word retrieval from weakly constraining contexts. Matched base/instruct variants provide a controlled within-family contrast, holding architecture and pretraining family as constant as public checkpoints permit. For each experiment, \textbf{we conduct matched human experiments with identical stimuli and design}.

Across both tasks, we find a consistent dissociation. Base models show strong, immediate facilitation that is stable across lags and robust to intervening material. Instruct models show weaker facilitation that decays with lag and, at larger scales, reverses to interference---the same word that helped the model earlier now hurts it. This is not a subtle quantitative difference: it is a qualitative split between two processing modes, suggesting that instruction tuning does not merely reshape \emph{what} models output but fundamentally changes \emph{how} they process repeated information.

Three mechanistic probes support this interpretation. Removing prior response context produces residual facilitation in base models but interference in instruct models; remapping classification labels shows partial transfer in base models but none in instruct models; and ablating attention heads associated with prior-occurrence retrieval sharply reduces priming only in base models. Together, these results indicate that base-model priming is at least partially word-level and stimulus-driven, while instruct-model priming depends more on downstream contextual evaluation. Within the controlled Qwen~2.5 series (1.5B--14B), the dissociation widens monotonically with scale. Matched human experiments reveal a third profile: lag-sensitive like instruct models but uniformly facilitative like base models, never reversing to interference.

\section{Related Work}

\subsection{In-Context Learning and Repetition}

In-context learning (ICL) enables models to adapt behavior based on prompt examples without parameter updates \citep{brown2020language}, with mechanistic studies identifying attention-based retrieval processes such as induction heads \citep{olsson2022context} and characterizing ICL as implicit optimization \citep{dai2023gpt, akyurek2023}. ICL research focuses on \emph{task-level} adaptation from structured examples; we study \emph{item-level} repetition across intervening material. Lag manipulation distinguishes durable item-specific effects from short-range recency phenomena \citep{min2022rethinking, zhao2021, liu2022}. Relatedly, \citet{vaidya2023humans} find that humans and LMs diverge when predicting repeating text, linking LM repetition to attention heads and recency. Our controlled design isolates the processing consequence of a repeated lexical item and compares matched base/instruct checkpoints.

\subsection{Priming in Language Models}

\emph{Semantic priming} studies report partial correspondence between human and model facilitation for related words \citep{ettinger2016probing, misra2020exploring}, while \emph{structural priming} studies show reuse of recent syntactic constructions \citep{sinclair2022structural}. Repetition priming instead concerns the \emph{same} word. \citet{mahaut2025repetitions} show that natural and ICL-induced repetition in generation rely on distinct mechanisms, differing in confidence, attention allocation, and sensitivity to perturbation. Their task concerns generated repetition; ours concerns controlled repeated-item processing. Both results show that superficially similar repetition effects can arise from distinct mechanisms. We use the diagnostic toolkit of human repetition-priming research and the automatic vs.\ controlled framework \citep{shiffrin1977controlled} to characterize how post-training shifts the default processing mode.

\subsection{Effects of Instruction Tuning}

Instruction tuning is known to reshape model behavior beyond surface-level compliance \citep{ouyang2022training}. \citet{wu2024behavior} show that instruction tuning adjusts self-attention heads to capture instruction-relevant relationships and rotates feed-forward representations toward user-oriented tasks, suggesting deep representational changes rather than superficial formatting shifts. Comparisons of base and instruct variants reveal systematic behavioral differences: base models outperform instruct models in retrieval-augmented generation settings \citep{cuconasu2024tale}, instruction-tuned models show improved knowledge acquisition \citep{jiang2024instruction}, while instruct models show enhanced but also more brittle task-following behavior. These findings motivate our question: does instruction tuning also change how models process repeated information at the item level? We provide evidence that it does, shifting the default from automatic to controlled processing. Extended related work on dual-process frameworks, LLM--human cognitive comparisons, and repetition suppression in neuroscience appears in Appendix~\ref{app:related}.

\section{Experiment 1: Semantic Categorization}
\label{sec:exp1}

\subsection{Models}

We tested 15 models spanning 5 families and 5 parameter scales (Table~\ref{tab:models}). At the 7--8B scale, we tested LLaMA~3 (8B), Mistral (7B), and Qwen~2.5 (7B), each with base and instruct variants. We also tested Gemma~2 (2B, base and instruct) and Phi-4 (14B, instruct only). To investigate scaling, we tested Qwen~2.5 at four scales: 1.5B, 3B, 7B, and 14B (base and instruct at each).

All models were run in bfloat16 precision with eager attention for attention matrix extraction.

\begin{table}[tb]
\centering
\small
\begin{tabular}{llr}
\toprule
\textbf{Model Family} & \textbf{Variants} & \textbf{Params} \\
\midrule
LLaMA 3 & base, instruct & 8B \\
Mistral & base, instruct & 7B \\
Gemma 2 & base, instruct & 2B \\
Phi-4 & instruct & 14B \\
\midrule
\multirow{4}{*}{Qwen 2.5} & base, instruct & 1.5B \\
 & base, instruct & 3B \\
 & base, instruct & 7B \\
 & base, instruct & 14B \\
\bottomrule
\end{tabular}
\caption{Models tested (15 models across 5 families). The Qwen~2.5 family spans four parameter scales, enabling analysis of how model size interacts with instruction tuning to affect information processing.}
\label{tab:models}
\end{table}

\subsection{Stimuli and Design}

We selected 182 concrete nouns from the \citet{mcrae2005semantic} feature norms, balanced between living entities (91) and nonliving objects (91). Forty-five non-target concrete nouns served as fillers, balanced across category (22 living, 23 nonliving); fillers did not repeat within a target's intervening window. A control for the proportion of same-category fillers did not predict target priming ($\beta=0.02$, $SE=0.05$, $t=0.41$, $p=.68$) or alter the exposure and lag effects.

Each target word appeared 4 times (exposures E1--E4), with a fixed number of filler trials between consecutive exposures determined by the lag condition (1, 3, 7, or 15 intervening trials). Each lag condition was run as a separate session with context reset before each target word's first exposure, ensuring that priming effects reflect only within-word repetition. This within-item design allows us to measure the change in processing from E1 (baseline) to subsequent exposures.

\subsection{Procedure and Measures}

On each trial, models received a zero-shot classification prompt and were asked to categorize a word as ``living'' or ``nonliving.'' Following \citet{petroni2019language}, we measured the log-probability margin:
\begin{equation}
\text{Margin} = \log P(\text{correct}) - \log P(\text{incorrect})
\end{equation}
The priming effect, our key measure of how repeated exposure changes information processing, was computed as:
\begin{equation}
\text{Priming}_{E_i} = \text{Margin}_{E_i} - \text{Margin}_{E_1}
\end{equation}
Positive values indicate facilitation; negative values indicate interference.

\subsection{Experiment 1a: LLM Results}

\begin{figure}[t]
    \centering
    \includegraphics[width=\textwidth]{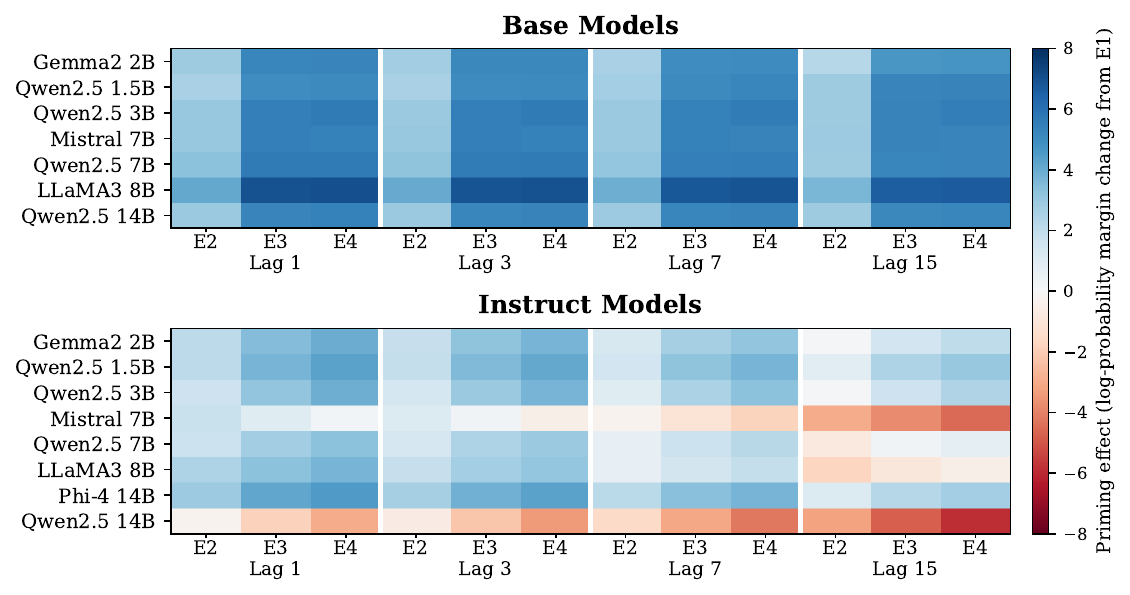}
    \caption{Experiment~1 priming profiles across lag conditions and exposures. Base and instruct models are shown in separate enlarged panels; blue indicates facilitation and red indicates interference. Numerical exposure means and fitted lag coefficients are reported in Appendix Tables~\ref{tab:exp1_overall} and~\ref{tab:exp1_mixed}.}
    \label{fig:exp1_heatmap}
\end{figure}

\paragraph{Priming magnitude.} All 7 base models showed strong, uniformly positive priming by the final exposure (E4$-$E1 range: $+5.04$ to $+6.89$, all $p<.001$; Table~\ref{tab:exp1_overall}; Figure~\ref{fig:exp1_heatmap}). Instruct models were weaker and more variable: most showed moderate facilitation, but two showed interference (Mistral-Instruct and Qwen2.5-14B-Instruct; see Table~\ref{tab:exp1_overall}).

\paragraph{Lag profiles.} To estimate the joint effects of exposure number and lag on priming magnitude, we fit a linear mixed-effects model separately for each model.\footnote{Per-model statistics are reported for transparency; our claims rest on the consistency of the qualitative pattern across model families, not on individual $p$-values.}
\begin{equation}
 \text{M}_{ij} = \beta_0 + \beta_1 \text{Exp}_i + \beta_2 \text{Lag}_j + (1 + \text{Exp}_i \mid \text{Word}) + \varepsilon_{ij}
\label{eq:mixed}
\end{equation}
where $\text{M}_{ij}$ is the log-probability margin, $\text{Exp}_i$ is ordinal exposure, $\text{Lag}_j$ is the number of intervening items, and $(1 + \text{Exp}_i \mid \text{Word})$ allows baseline difficulty and repetition sensitivity to vary by item. All 7 base models showed positive exposure effects (all $p<.001$) and no lag effects. Six of eight instruct models showed negative lag effects at uncorrected thresholds; five remained significant after FDR correction. Facilitation therefore decayed with lag and, in Qwen2.5-14B-Instruct, reversed to interference (Table~\ref{tab:exp1_mixed}).

\begin{figure}[t]
    \centering
    \includegraphics[width=0.394\textwidth]{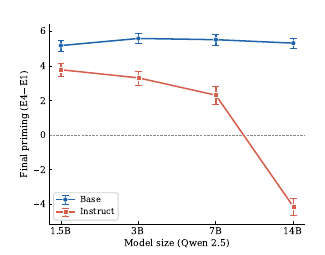}
    \caption{Experiment~1 scaling trends (Qwen~2.5). Final priming (E4$-$E1) is stable for base but declines for instruct models. Error bars: SEM.}
    \label{fig:exp1_scaling}
\end{figure}

\paragraph{Scaling trend.} The Qwen~2.5 family (1.5B, 3B, 7B, 14B) provides the only continuous matched size series in our model set (Figure~\ref{fig:exp1_scaling}). Within this family, base priming was stable across scale, while instruct priming declined monotonically and reversed to interference at 14B. We do not infer from this within-Qwen pattern that size alone explains cross-family variation.

\subsection{Experiment 1b: Human Experiment}
\label{sec:exp1b}

To anchor the LLM findings in human cognition, we conducted a matched semantic categorization experiment with 40 native English speakers (22 female, 18 male; mean age = 23.4, $SD = 3.1$). Humans classified a subset of 10 target words (5 living, 5 nonliving) using the same lag conditions and per-word context isolation design, with response time (RT) and accuracy as dependent measures.

\paragraph{Human priming: facilitative and lag-sensitive.} Humans showed robust facilitation in both RT and accuracy. Mean RT decreased from 648\,ms at E1 to 539\,ms at E4 ($t(39) = 11.82$, $p < .001$, $d = 1.87$), and accuracy improved from 92.3\% at E1 to 97.1\% at E4 ($t(39) = 3.28$, $p = .002$, $d = 0.52$). Critically, human priming showed lag-dependent decay in RT ($\beta = -1.72$, $t = -5.84$, $p < .001$), well-described by a power-law function consistent with \citet{mckone1995}. Accuracy priming was less lag-sensitive ($\beta = -0.08$, $t = -1.42$, $p = .16$), likely reflecting a ceiling effect. Humans also showed a significant low-frequency advantage in RT (131\,ms vs.\ 87\,ms, $p = .003$).

\paragraph{Comparison with LLM patterns.} Human lag sensitivity resembles the instruct model pattern, not the base model's lag invariance. However, human priming was uniformly facilitative (never producing interference), unlike the interference observed in some instruct models. The low-frequency advantage in humans also contrasts with the null frequency effect in LLMs. We return to these convergences and divergences in the Discussion (Section~\ref{sec:discussion}).

\section{Experiment 2: Cloze Completion}
\label{sec:exp2}

Experiment~1 established the processing mode dissociation in forced-choice classification. Experiment~2 tests whether the same dissociation extends to open-ended word retrieval, where models must recover a specific lexical item from weakly constraining sentence contexts, providing a purer test of how repeated exposure changes information retrieval.

\subsection{Stimuli}

We used standardized completion norms from \citet{peelle2020completion} (3085 English sentence contexts normed by 200+ participants). We selected 21 target words, each paired with 4 low-cloze sentence frames (cloze probability: 0.05--0.15). At these cloze levels, the sentence context is weakly constraining: many words are plausible completions, so no single word dominates. For example, ``James had to see his doctor about his \underline{\hspace{1cm}}'' (target: \textit{rash}, cloze = 0.09) admits \textit{back}, \textit{leg}, \textit{cough}, etc.; ``Greg looked down and realized he was missing his \underline{\hspace{1cm}}'' (target: \textit{ring}, cloze = 0.15) admits \textit{wallet}, \textit{keys}, \textit{phone}, etc. Using exclusively low-cloze sentences ensures that the sentence alone weakly constrains the target, so improved performance can only arise from prior in-context exposure, making this a direct test of information retrieval mechanisms.

Target words satisfied three constraints: (1)~$\geq$4 low-cloze sentences available, (2)~single-word completions, and (3)~zero contamination, meaning the target word does not appear in any other sentence. The 21 words comprised 14 nouns (\textit{church}, \textit{fear}, \textit{friend}, \textit{horse}, \textit{mud}, \textit{names}, \textit{nose}, \textit{rash}, \textit{rat}, \textit{ring}, \textit{salt}, \textit{song}, \textit{speed}, \textit{spider}), 6 verbs (\textit{fight}, \textit{laugh}, \textit{lure}, \textit{sail}, \textit{steal}, \textit{wash}), and 1 adjective (\textit{sour}); full list in Appendix~\ref{app:stimuli}. We selected 50 filler sentence--word pairs verified to contain no target word; the same fillers were reused across lag conditions.

\subsection{Design and Procedure}

Each target word appeared 4 times (E0--E3) across different sentence frames, with 1, 3, or 7 filler trials between exposures. Context was reset before each word's first exposure (E0), ensuring that priming effects reflect only within-word repetition. A baseline session presented each sentence in isolation.

\subsection{Analysis}

The dependent measure parallels Experiment~1: we track how the log-probability assigned to the target word changes across repeated exposures. Because cloze completion is open-ended (no fixed incorrect alternative), we use the raw log-probability of the target rather than a margin between correct and incorrect options.\footnote{Raw log-probabilities may partly reflect sentence-level entropy differences across frames. Our within-item design (each word serves as its own baseline at E0) controls for this, since priming is computed as a within-word change. We verified that results are qualitatively unchanged when using rank-normalized probabilities.} Priming is defined as:
\begin{equation}
\text{Priming}_{E_i} = \log P(\text{target})_{E_i} - \log P(\text{target})_{E_0}
\end{equation}
where $E_0$ is the first in-session exposure. We use E0 rather than the isolated baseline session as the reference because E0 reflects the model's response to the target word within the same context window and session format as subsequent exposures, providing a cleaner within-session comparison. The isolated baseline session serves as a verification that E0 performance is comparable to truly naive performance (which it is: mean difference $<0.05$ log-probability units across models). We report priming at each subsequent exposure (E1$-$E0, E2$-$E0, E3$-$E0) and fit the same mixed-effects model structure as Experiment~1 (Equation~\ref{eq:mixed}), replacing margin with log-probability as the dependent variable and using lag $\in \{1,3,7\}$.

\subsection{Experiment 2a: LLM Results}

\begin{figure}[t]
    \centering
    \begin{subfigure}[t]{0.48\columnwidth}
        \centering
        \includegraphics[width=\textwidth]{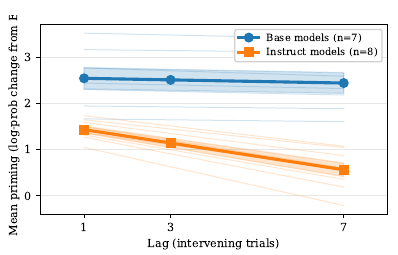}
        \caption{Mean priming by lag. Base models (blue) maintain stable priming; instruct models (orange) decline with lag.}
        \label{fig:exp2_learning}
    \end{subfigure}
    \hfill
    \begin{subfigure}[t]{0.48\columnwidth}
        \centering
        \includegraphics[width=\textwidth]{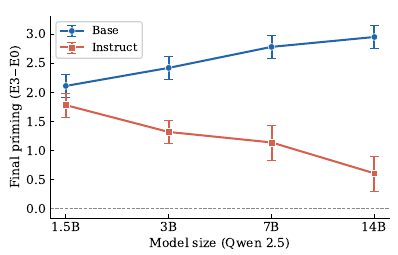}
        \caption{Scaling trends (Qwen~2.5). Final priming (E3$-$E0) improves with scale for base models but declines for instruct models.}
        \label{fig:exp2_scaling}
    \end{subfigure}
    \caption{Experiment~2 results. (a) Thin lines: individual models; thick lines: group means. Error bars/shaded regions: SEM. (b) Error bars: SEM.}
    \label{fig:exp2_combined}
\end{figure}

\paragraph{Priming magnitude.} All 7 base models showed strong positive priming (E3$-$E0 range: $+2.11$ to $+4.12$, all $p<.001$; Table~\ref{tab:exp2_overall}; Figure~\ref{fig:exp2_learning}). Instruct models showed consistently weaker effects (E3$-$E0 range: $+0.61$ to $+1.85$), with the base/instruct gap present across all families.

\paragraph{Lag profiles replicate the Experiment~1 dissociation.} All base models showed significant exposure effects with no lag sensitivity, while all eight instruct models showed negative lag effects at uncorrected thresholds; six remained significant after FDR correction (Table~\ref{tab:exp2_mixed}). Omnibus mixed-effects models confirmed ModelType$\times$Lag interactions in Experiment~1 ($\beta=-0.26$, $SE=0.04$, $t=-6.50$, $p<.001$) and Experiment~2 ($\beta=-0.13$, $SE=0.03$, $t=-4.52$, $p<.001$). Main between-type effect sizes ranged from $d=1.82$ to $2.93$ (Appendix~\ref{app:stats}).

\paragraph{Prompt robustness.} Re-running Experiment~1 with three semantically equivalent instruction/answer-cue templates preserved the LLaMA3 base/instruct split: base lag slopes ranged from $-0.04$ to $+0.01$ (all n.s.), versus $-0.32$ to $-0.27$ for instruct (all $p<.001$); the Prompt$\times$ModelType$\times$Lag interaction was not significant ($F=1.12$, $p=.33$). Analysis details appear in Appendix~\ref{app:prompts}.

\paragraph{Scaling trend.} Within the Qwen~2.5 family, base model priming improved with scale while instruct model priming declined (Figure~\ref{fig:exp2_scaling}), replicating the Experiment~1 scaling pattern.

\subsection{Experiment 2b: Human Experiment}
\label{sec:exp2b}

Another 38 native English speakers (20 female, 18 male; mean age = 22.8, $SD = 2.9$) completed a matched cloze experiment. On each trial, participants saw a sentence frame with a blank and typed a single-word completion. In the \emph{study phase}, participants read a complete sentence containing the target word (e.g., ``The man was professionally trained in how to \textbf{fight}.''). In the \emph{test phase}, after 1, 3, or 7 filler trials, participants completed a different sentence frame for the same target word (e.g., ``Afraid of getting injured, Zach backed out of the \_\_\_\_\_.''). Dependent measures were \emph{target production rate} (whether the participant produced the target word) and \emph{response time} (RT). A baseline condition presented test sentences without any prior study exposure.

\paragraph{Human priming: gradual and lag-sensitive.} Prior exposure significantly increased target production rate: from 11.2\% at baseline to 28.4\% after one study exposure ($\chi^2(1) = 18.7$, $p < .001$), rising to 41.7\% after three exposures. RT for correctly produced targets also decreased with exposure (from 4.82\,s at E1 to 3.41\,s at E3; $t(37) = 4.93$, $p < .001$, $d = 0.78$). Both measures showed significant lag-dependent decay (production rate: $\beta = -2.1\%$ per lag unit, $z = -3.42$, $p < .001$; RT: $\beta = +0.18$\,s per lag unit, $t = 2.87$, $p < .01$). This pattern resembles the instruct model trajectory: gradual, lag-sensitive improvement rather than the base model's near-instantaneous retrieval.

\paragraph{Comparison with LLM patterns.} As in Experiment~1b, human priming was lag-sensitive (resembling instruct models) but uniformly facilitative. Even at the longest lag, human target production remained above baseline (19.8\% vs.\ 11.2\%, $p < .01$). RT and model log-probability are not directly commensurate; we compare only the profile---facilitation versus interference, lag sensitivity, and reversal (Appendix Table~\ref{tab:profile}).

\section{Mechanistic Evidence}

The preceding experiments establish \emph{that} base and instruct models differ; the following analyses probe \emph{how}, by selectively disrupting contextual information and internal retrieval pathways.

\subsection{Evidence 1: Missing Context Reveals Opposite Processing Defaults}
\label{sec:ablation}

What happens when models encounter a repeated word but the prior response is missing? We tested this in Experiment~1 by replacing classification responses (``living''/``nonliving'') in the context history with placeholder tokens (``...'') while maintaining word repetition. (This ablation is not applicable to Experiment~2, where removing the target word completion would eliminate the word's presence in context entirely, precluding any repetition effect.)

\paragraph{Results.} Base models showed reduced but still positive priming (mean: $+0.70$; range: $+0.41$ to $+1.38$), approximately 15\% of the full-context effect. The repeated word still triggered partial activation even without response history, consistent with stimulus-driven processing. Seven of eight instruct models shifted to negative priming (range: $-1.26$ to $-4.84$; overall instruct mean: $-2.36$). The exception, Qwen2.5-14B-Instruct, showed positive priming at lag~1 that reversed at longer lags (mean $-0.61$; see Appendix Table~\ref{tab:ablation_lag}). Without the expected response context, controlled processing generates interference: the model detects the repeated word but interprets the incomplete context as evidence against its prior assessment.

This asymmetric response to missing context distinguishes the two processing modes. The same manipulation produces weakened facilitation from the automatic system (the stimulus alone still partially activates its representation) versus active interference from the controlled system (expected context missing, generating conflict).

\paragraph{Converging evidence from label remapping.} We reversed the classification labels after Phase~1 training (living$\rightarrow$A became living$\rightarrow$B) and tested whether priming survived. Among base models, 3 of 7 showed significant facilitation despite reversed labels, with the remaining four showing non-significant positive trends. Among instruct models, none showed significant transfer. This confirms that base model priming is at least partially word-level, while instruct model priming is tied to the specific stimulus-response context (details in Appendix~\ref{app:results}).

\begin{figure}[t]
    \centering
    \includegraphics[width=\textwidth]{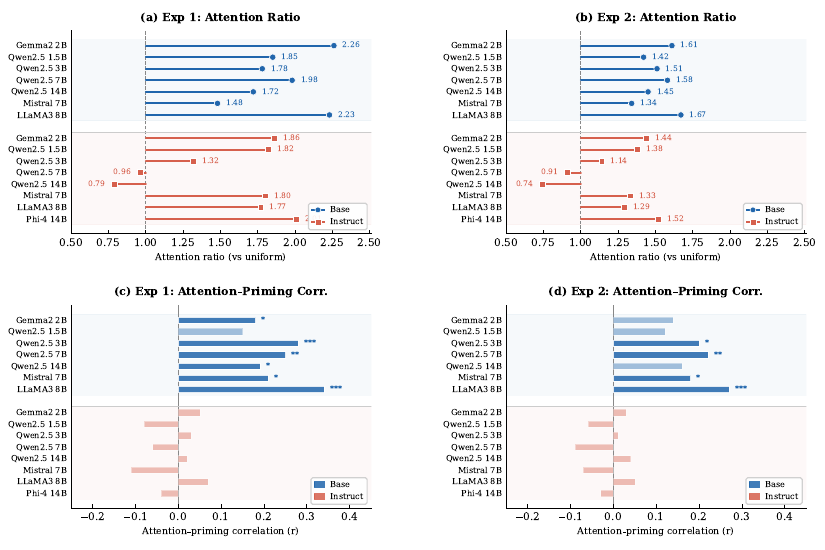}
    \caption{Attention analysis. \textbf{(a--b)} Attention ratio to prior occurrences vs.\ uniform baseline (dashed line). Base models show elevated attention; instruct models decline with scale. \textbf{(c--d)} Attention--priming correlation. Base models show significant positive correlations; instruct models show weak correlations.}
    \label{fig:attention}
\end{figure}

\subsection{Evidence 2: Attention Patterns Suggest Shared Retrieval with Divergent Downstream Processing}
\label{sec:attention}

The preceding evidence establishes \emph{behavioral} differences. Does the underlying retrieval mechanism also differ, or do both model types retrieve prior occurrences but process them differently? We examine this question through attention weight analysis across both experiments. We measured attention to positions where the target word had previously appeared. The \emph{attention ratio}, defined as attention to prior occurrences divided by the uniform-distribution expectation, quantifies whether models preferentially attend to prior context (Figure~\ref{fig:attention}, panels a--b).
\paragraph{Both model types attend to prior occurrences.} Across both experiments, base models showed consistently elevated attention ratios, while instruct models were more variable (Figure~\ref{fig:attention}, panels a--b). Ratios were lower in Experiment~2, consistent with the longer cloze contexts distributing attention more broadly. Most models attended to prior occurrences above baseline, suggesting that attention-based retrieval is a \emph{shared} mechanism.
\paragraph{The divergence appears downstream.} The attention--behavior relationship differed between model types (Figure~\ref{fig:attention}, panels c--d). Base models showed significant positive correlations between attention to prior occurrences and priming magnitude in both experiments, indicating that retrieval directly drives facilitation. Instruct models showed uniformly weak and non-significant correlations ($|r|<0.12$), suggesting that attention-based retrieval is decoupled from the downstream processing that determines the behavioral outcome.
\paragraph{Scaling trend.} Within the Qwen2.5 instruct family, attention ratio declined monotonically with scale across both tasks, with the largest models allocating \emph{less} attention to prior occurrences than expected by chance. This suggests that at sufficient scale, the post-training pipeline may begin to suppress attention-based retrieval itself.

\subsection{Evidence 3: Targeted Attention-Head Ablation}
We ranked attention heads by how strongly their attention to prior target occurrences predicted item-level priming, then ablated the top 10 during inference. Base-model priming dropped sharply (LLaMA3: $+6.89\rightarrow+2.62$; Qwen2.5: $+5.52\rightarrow+1.84$), whereas random-head controls largely preserved it ($+6.12$ and $+5.21$). The same targeted intervention barely changed instruct models (LLaMA3: $+2.05\rightarrow+1.97$; Qwen2.5: $+2.31\rightarrow+2.14$). Thus prior-occurrence attention is causally involved in base-model facilitation, whereas instruct behavior is comparatively insensitive to this pathway (Appendix Table~\ref{tab:head-ablation}).
\section{Discussion and Conclusion}
\label{sec:discussion}

The results map onto the \emph{automatic} vs.\ \emph{controlled processing} framework \citep{shiffrin1977controlled}. Base models exhibit \textbf{automatic processing}: repeated words trigger lag-invariant facilitation that partly survives context removal and label remapping. Instruct models exhibit \textbf{controlled processing}: context-dependent re-evaluation that decays with lag and collapses without expected context. Correlational attention results and causal head ablations converge: prior-occurrence retrieval directly supports base-model priming, but is filtered or gated before shaping instruct outputs. The source of that gating remains open. It may reflect conflict resolution or broader post-training-induced repetition avoidance, not necessarily RLHF alone; isolating SFT from preference optimization requires checkpoints along a controlled post-training trajectory. Human processing is lag-sensitive like instruct models but uniformly facilitative like base models \citep{bowers2000, tenpenny1995}. Practically, RAG, few-shot prompting, and dialogue all reuse contextual information; our results predict that the effects of repeated chunks, examples, or entities depend on model type and recency. Repetition priming therefore reveals a consequential change associated with post-training while also identifying a hybrid human profile that neither model type captures.

\section*{Ethics Statement}

The human experiments in this study were approved by the Institutional Review Board (IRB) of Yale University. All participants provided written informed consent prior to participation and were compensated at a rate of \$15/hour. The experiments involved only standard psycholinguistic tasks (word categorization and sentence completion) with no deceptive procedures, sensitive content, or foreseeable risks beyond those of everyday computer use. All human data were collected and stored in accordance with institutional data protection guidelines and are anonymized in all analyses.

\section*{LLM Usage Disclosure}
An LLM-based writing assistant was used to support language editing and camera-ready organization. All scientific claims, analyses, and final text were reviewed and verified by the authors.

\bibliographystyle{colm2026_conference}
\bibliography{references}

\appendix

\section{Extended Related Work}
\label{app:related}

\paragraph{Dual-Process Frameworks for LLMs.} The distinction between fast, heuristic processing (System~1) and slow, deliberate reasoning (System~2) \citep{kahneman2011thinking} has been increasingly applied to LLMs. \citet{brady2025dual} argue that LLMs exhibit both System-1-like biases and System-2-like reasoning depending on prompting strategy. \citet{li2025system} survey the rapid development of reasoning LLMs (e.g., OpenAI o1, DeepSeek R1), framing progress as a shift from System~1 to System~2 capabilities. \citet{ziabari2025reasoning} propose methods to align LLMs along this spectrum. However, these approaches primarily manipulate the \emph{prompting} or \emph{training} regime to elicit different reasoning modes. Our work differs in showing that the dual-process distinction emerges \emph{spontaneously} from instruction tuning, observable through a simple repetition priming paradigm without any explicit prompting for fast or slow reasoning.

\paragraph{LLM--Human Cognitive Comparisons.} A growing body of work evaluates LLMs as models of human cognition. \citet{sartori2023language} provide a broad assessment of language models as tools for psychological science, noting both striking correspondences and systematic divergences. \citet{duan2024hlb} benchmark LLMs on humanlikeness across multiple linguistic dimensions. In psycholinguistics specifically, LLMs replicate many human behavioral patterns including semantic priming and garden-path effects \citep{misra2020exploring, ettinger2016probing}, but also show non-human patterns such as insensitivity to word frequency effects that are robust in human data. Our matched human experiments contribute to this literature by identifying a specific phenomenon, repetition priming, where human behavior falls between two distinct LLM processing modes, rather than aligning with either.

\paragraph{Effects of Instruction Tuning on Internal Representations.} Beyond behavioral changes, instruction tuning reshapes model internals. \citet{wu2024behavior} demonstrate that instruction tuning modifies self-attention to capture instruction-verb relationships and rotates feed-forward knowledge toward task-relevant directions. Mechanistic interpretability studies show that fine-tuning surfaces pre-existing capabilities rather than constructing entirely new circuits \citep{olsson2022context}. Cross-model steering experiments suggest shared representational subspaces between base and instruct variants. Our attention analysis (Section~\ref{sec:attention}) complements these findings by showing that while both model types share an attention-based retrieval mechanism for prior occurrences, instruction tuning decouples this retrieval from downstream behavioral effects.

\paragraph{Repetition Suppression in Neuroscience.} In cognitive neuroscience, repeated stimulus presentation typically produces \emph{repetition suppression} (reduced neural responses to familiar stimuli) alongside behavioral facilitation \citep{henson2003, henson2004}. This dissociation between neural suppression and behavioral facilitation parallels our finding that instruct models can show reduced attention to prior occurrences (especially at scale) while still exhibiting some behavioral facilitation. The neuroscience literature also emphasizes that repetition effects are modulated by attention, expectation, and recognition \citep{grill2006repetition}, variables that map onto the controlled processing characteristics we observe in instruct models.

\paragraph{Distinct Repetition Mechanisms.} \citet{vaidya2023humans} show that human and LM predictions diverge for repeating passages and relate the LM pattern to middle-layer attention heads and recency. \citet{mahaut2025repetitions} further distinguish natural repetition from ICL-induced repetition: the two patterns differ in confidence, specialized heads, and sensitivity to perturbations. Our task differs from both by measuring controlled repetition priming rather than repeated generation, but the conceptual convergence is important: repetition is not a unitary mechanism, and superficially similar effects can depend on different attention pathways and contextual demands.

\section{Experimental Design Details}
\label{app:design}

\paragraph{Lag Manipulation.} In both experiments, we manipulated the number of intervening trials between successive presentations of a target word. Table~\ref{tab:design} summarizes the key design parameters. For Experiment 1, lag values of 1, 3, 7, and 15 were tested; for Experiment 2, lag values of 1, 3, and 7 were tested, plus an isolated baseline session. Each lag condition was run as a separate session with fresh context.

\paragraph{Trial Structure.} In both experiments, each target word was tested in an independent context block: context was reset before each word's first exposure, ensuring priming effects reflect only within-word repetition. In Experiment 1, each trial presented a single word for classification, with 45 filler words providing intervening trials. In Experiment 2, each trial presented a sentence with a blank to be completed.

\begin{table}[H]
\centering
\footnotesize
\begin{tabular}{lcc}
\toprule
\textbf{Parameter} & \textbf{Exp 1} & \textbf{Exp 2} \\
\midrule
Target words & 182 & 21 \\
Fillers & 45 & 50 \\
Presentations per word & 4 (E1--E4) & 4 (E0--E3) \\
Lag conditions & 1, 3, 7, 15 & 1, 3, 7 \\
Cloze range & --- & 0.05--0.15 \\
Context isolation & per word & per word \\
Sessions per model & 728 & 84 \\
\bottomrule
\end{tabular}
\caption{Experimental design parameters.}
\label{tab:design}
\end{table}

\section{Stimuli}
\label{app:stimuli}

\paragraph{Experiment 1.} 182 target words were selected from the McRae feature norms (91 living, 91 nonliving). Table~\ref{tab:exp1_words} shows representative examples.

\begin{table}[H]
\centering
\footnotesize
\begin{tabular}{lp{5.5cm}}
\toprule
\textbf{Category} & \textbf{Examples} \\
\midrule
Living (91) & tiger, elephant, artist, baby, child, robin, salmon, \ldots \\
Nonliving (91) & bus, car, bicycle, bowl, jacket, hammer, violin, \ldots \\
\bottomrule
\end{tabular}
\caption{Target words for Experiment 1 (182 words total, selected from McRae feature norms).}
\label{tab:exp1_words}
\end{table}

\paragraph{Experiment 2.} The 21 target words were: \textit{church, fear, fight, friend, horse, laugh, lure, mud, names, nose, rash, rat, ring, sail, salt, song, sour, speed, spider, steal, wash}. Each was paired with 4 low-cloze sentence frames (cloze 0.05--0.15). Example for \textit{fight}: (E0) ``The man was professionally trained in how to \_\_\_\_\_.'' (cloze = 0.13), (E1) ``Afraid of getting injured, Zach backed out of the \_\_\_\_\_.'' (cloze = 0.14), (E2) ``The boys were told not to \_\_\_\_\_.'' (cloze = 0.08), (E3) ``They ended up having a huge \_\_\_\_\_.'' (cloze = 0.11).

\section{Prompt Formats}
\label{app:prompts}

For Experiment~1, the prompt-sensitivity analysis used the following three semantically equivalent templates:

\begin{quote}
\small
\textbf{Template 1 (original)}\\
\texttt{Task: Classify each word as `living' or `nonliving'.}\\
\texttt{Word: tiger}\\
\texttt{Answer: living}\\
\texttt{Word: bus}\\
\texttt{Answer:}\\[0.6em]
\textbf{Template 2}\\
\texttt{Decide whether each word denotes a living thing or a nonliving thing.}\\
\texttt{Word: tiger}\\
\texttt{Category: living}\\
\texttt{Word: bus}\\
\texttt{Category:}\\[0.6em]
\textbf{Template 3}\\
\texttt{For each item, choose exactly one label: living or nonliving.}\\
\texttt{Item: tiger}\\
\texttt{Label: living}\\
\texttt{Item: bus}\\
\texttt{Label:}
\end{quote}

\noindent Instruct models received each version wrapped in the appropriate chat template; response labels, trial history, item order, and scored tokens were otherwise unchanged. For Experiment~2: \texttt{Complete the sentence with one word: The group of friends would often \_\_\_. Answer: laugh}

\paragraph{Prompt-sensitivity control.} We evaluated three semantically equivalent Experiment~1 templates that varied the instruction wording and answer cue while preserving the response options. Across templates, the LLaMA3-Base lag coefficient ranged from $-0.04$ to $+0.01$ (all n.s.), whereas LLaMA3-Instruct ranged from $-0.32$ to $-0.27$ (all $p<.001$). The Prompt$\times$ModelType$\times$Lag interaction was not significant ($F=1.12$, $p=.33$), indicating that wording differences did not explain the dissociation.

\paragraph{Ablation Condition.} Responses replaced with ``...'' placeholders.

\paragraph{Label Remapping.} Abstract labels (A/B) instead of semantic labels. Phase~1: 20 words $\times$ 3 exposures with M1. Mapping change + 8 warmup trials. Test: 20 TARGET vs.\ 20 CONTROL words under M2.

\section{Model Specifications}
\label{app:models}

All experiments used HuggingFace Transformers (v4.36.0) with bfloat16 precision, eager attention, greedy decoding, and 100-trial context limit. Hardware: NVIDIA A100 80GB GPUs.

\section{Statistical Analysis Details}
\label{app:stats}

\paragraph{Priming Effect.} $\text{Priming}_{E_i} = \text{Margin}_{E_i} - \text{Margin}_{E_1}$. We report priming separately for each exposure (E2$-$E1, E3$-$E1, E4$-$E1) as well as the mean across E2--E4, averaged across all target words and lag conditions.

\paragraph{Omnibus models and multiplicity.} Within each experiment, an omnibus mixed-effects model included ModelType, Exposure, Lag, and their interactions as fixed effects, with random intercepts for Model and Word. The ModelType$\times$Lag interaction was significant in Experiment~1 ($\beta=-0.26$, $SE=0.04$, $t=-6.50$, $p<.001$) and Experiment~2 ($\beta=-0.13$, $SE=0.03$, $t=-4.52$, $p<.001$). False-discovery-rate correction preserved negative lag effects for 5/8 instruct models in Experiment~1 and 6/8 in Experiment~2, versus 0/7 base models in either experiment.

\paragraph{Effect sizes.} Standardized between-type contrasts were large: Experiment~1 mean priming $d=2.18$, Experiment~1 final priming $d=1.82$, Experiment~2 mean priming $d=2.90$, and the missing-context contrast $d=2.93$. In label remapping, LLaMA3-Base retained a $+1.00$ TARGET-over-CONTROL margin after remapping ($p<.01$, $d=0.87$).

\paragraph{Attention Analysis.} Attention to prior occurrences: (1) identify positions of prior presentations, (2) sum attention weights from final token, (3) average across heads per layer, (4) compute ratio to uniform-distribution expectation.

\paragraph{Targeted head ablation.} For LLaMA3-8B and Qwen2.5-7B base/instruct pairs, we ranked individual heads by the association between attention to earlier target occurrences and item-level priming, selected the top 10, and ablated their outputs during inference. Random sets of 10 heads served as controls. Table~\ref{tab:head-ablation} reports final priming under each condition.

\begin{table}[H]
\centering
\footnotesize
\begin{tabular}{lrrr}
\toprule
\textbf{Model} & \textbf{Full} & \textbf{Top-10} & \textbf{Random-10} \\
\midrule
LLaMA3 Base & +6.89 & +2.62 & +6.12 \\
LLaMA3 Instruct & +2.05 & +1.97 & +2.01 \\
Qwen2.5 Base & +5.52 & +1.84 & +5.21 \\
Qwen2.5 Instruct & +2.31 & +2.14 & +2.25 \\
\bottomrule
\end{tabular}
\caption{Targeted attention-head ablation in Experiment~1. Ablating heads most associated with prior-occurrence retrieval reduces base-model priming, whereas instruct models and random-head controls change little.}
\label{tab:head-ablation}
\end{table}

\section{Human--Model Comparison Details}
\label{app:human-comparison}

The human and model dependent measures have different units and should not be compared in absolute magnitude. Human RT/accuracy/production serve as a cognitive reference for the qualitative shape of the effect.

\begin{table}[H]
\centering
\footnotesize
\begin{tabular}{lccc}
\toprule
\textbf{Pattern} & \textbf{Base LMs} & \textbf{Instruct LMs} & \textbf{Humans} \\
\midrule
Facilitation & Strong positive & Weaker/variable & Positive RT/accuracy \\
Lag sensitivity & None & Predominantly negative & Significant in RT \\
Long-lag interference & Not observed & Some models & Not observed \\
\bottomrule
\end{tabular}
\caption{Profile-level comparison. Human behavioral measures and model log-probabilities are used to compare direction and lag dependence, not metric-level magnitudes.}
\label{tab:profile}
\end{table}

\begin{table}[H]
\centering
\footnotesize
\begin{tabular}{llrr}
\toprule
\textbf{Experiment} & \textbf{Measure} & \textbf{Initial/baseline} & \textbf{Final} \\
\midrule
Semantic categorization & RT & 648 ms & 539 ms \\
 & Accuracy & 92.3\% & 97.1\% \\
Cloze completion & Target production & 11.2\% & 41.7\% \\
 & Correct-production RT & 4.82 s & 3.41 s \\
\bottomrule
\end{tabular}
\caption{Human performance summaries. Experiment~1 compares E1 with E4. Cloze production compares the no-study baseline with three exposures; RT compares E1 with E3 among correctly produced targets.}
\label{tab:human-summary}
\end{table}

\section{Additional Results}
\label{app:results}

\begin{table}[H]
\centering
\footnotesize

\begin{tabular}{llcccc}
\toprule
\textbf{Model} & \textbf{Size} & \textbf{E2$-$E1} & \textbf{E3$-$E1} & \textbf{E4$-$E1} & \textbf{Mean} \\
\midrule
\multicolumn{6}{l}{\textit{Base models}} \\
LLaMA3 & 8B & +3.94 & +6.82 & +6.89 & +5.88 \\
Qwen2.5-7B & 7B & +3.12 & +5.50 & +5.52 & +4.71 \\
Qwen2.5-3B & 3B & +2.95 & +5.42 & +5.59 & +4.65 \\
Mistral & 7B & +2.99 & +5.42 & +5.34 & +4.58 \\
Qwen2.5-14B & 14B & +2.91 & +5.21 & +5.32 & +4.48 \\
Qwen2.5-1.5B & 1.5B & +2.68 & +5.10 & +5.18 & +4.32 \\
Gemma2 & 2B & +2.64 & +4.99 & +5.04 & +4.22 \\
\midrule
\multicolumn{6}{l}{\textit{Instruct models}} \\
Phi-4 & 14B & +2.21 & +3.41 & +3.80 & +3.14 \\
Qwen2.5-1.5B & 1.5B & +1.59 & +3.24 & +3.78 & +2.87 \\
Gemma2 & 2B & +1.34 & +2.72 & +3.21 & +2.42 \\
Qwen2.5-3B & 3B & +1.05 & +2.58 & +3.31 & +2.31 \\
Qwen2.5-7B & 7B & +0.73 & +1.82 & +2.31 & +1.62 \\
LLaMA3 & 8B & +0.80 & +1.64 & +2.05 & +1.50 \\
Mistral & 7B & $-$0.10 & $-$0.92 & $-$1.63 & $-$0.88 \\
Qwen2.5-14B & 14B & $-$1.41 & $-$2.98 & $-$4.14 & $-$2.84 \\
\bottomrule
\end{tabular}
\caption{Priming by exposure in Experiment 1 (change in log-probability margin from E1 baseline, averaged across lag conditions).}
\label{tab:exp1_overall}
\end{table}

\begin{table}[H]
\centering
\footnotesize

\begin{tabular}{llcccc}
\toprule
\textbf{Model} & \textbf{Size} & \multicolumn{2}{c}{\textbf{Exposure}} & \multicolumn{2}{c}{\textbf{Lag}} \\
\cmidrule(lr){3-4} \cmidrule(lr){5-6}
 & & $\beta$ & $t$ & $\beta$ & $t$ \\
\midrule
\multicolumn{6}{l}{\textit{Base models}} \\
Qwen2.5-3B & 3B & +1.62 & 11.24*** & $-$0.01 & $-$0.34 \\
LLaMA3 & 8B & +1.94 & 11.98*** & $-$0.03 & $-$1.07 \\
Qwen2.5-1.5B & 1.5B & +1.52 & 8.41*** & +0.02 & 0.87 \\
Qwen2.5-7B & 7B & +1.58 & 11.75*** & $-$0.03 & $-$1.08 \\
Mistral & 7B & +1.56 & 10.57*** & $-$0.01 & $-$0.28 \\
Gemma2 & 2B & +1.48 & 8.79*** & $-$0.04 & $-$1.28 \\
Qwen2.5-14B & 14B & +1.55 & 9.82*** & $-$0.01 & $-$0.42 \\
\midrule
\multicolumn{6}{l}{\textit{Instruct models}} \\
Qwen2.5-7B & 7B & +0.62 & 1.38 & $-$0.18 & $-$3.24** \\
Qwen2.5-3B & 3B & +0.88 & 1.72 & $-$0.11 & $-$1.92 \\
Gemma2 & 2B & +1.18 & 3.26** & $-$0.14 & $-$1.48 \\
Qwen2.5-1.5B & 1.5B & +1.08 & 2.82** & $-$0.09 & $-$2.10* \\
Phi-4 & 14B & +1.01 & 2.40* & $-$0.13 & $-$3.08** \\
LLaMA3 & 8B & +0.57 & 1.25 & $-$0.30 & $-$6.74*** \\
Mistral & 7B & $-$0.14 & $-$0.39 & $-$0.34 & $-$11.36*** \\
Qwen2.5-14B & 14B & $-$0.97 & $-$1.59 & $-$0.21 & $-$3.84*** \\
\bottomrule
\end{tabular}
\caption{Mixed-effects model results (Experiment 1). Exposure $\beta$: change per additional exposure. Lag $\beta$: change per unit lag increase. *$p<.05$, **$p<.01$, ***$p<.001$.}
\label{tab:exp1_mixed}
\end{table}

\begin{table}[H]
\centering
\footnotesize

\begin{tabular}{llcccc}
\toprule
\textbf{Model} & \textbf{Size} & \textbf{E1$-$E0} & \textbf{E2$-$E0} & \textbf{E3$-$E0} & \textbf{Mean} \\
\midrule
\multicolumn{6}{l}{\textit{Base models}} \\
LLaMA3 & 8B & +2.41 & +3.87 & +4.12 & +3.47 \\
Gemma2 & 2B & +2.18 & +3.52 & +3.71 & +3.14 \\
Mistral & 7B & +1.82 & +3.01 & +3.24 & +2.69 \\
Qwen2.5-14B & 14B & +1.54 & +2.68 & +2.95 & +2.39 \\
Qwen2.5-7B & 7B & +1.45 & +2.51 & +2.78 & +2.25 \\
Qwen2.5-3B & 3B & +1.21 & +2.14 & +2.42 & +1.92 \\
Qwen2.5-1.5B & 1.5B & +0.98 & +1.82 & +2.11 & +1.64 \\
\midrule
\multicolumn{6}{l}{\textit{Instruct models}} \\
Gemma2 & 2B & +0.92 & +1.54 & +1.85 & +1.44 \\
Qwen2.5-1.5B & 1.5B & +0.88 & +1.47 & +1.78 & +1.38 \\
Phi-4 & 14B & +0.81 & +1.38 & +1.62 & +1.27 \\
Mistral & 7B & +0.72 & +1.18 & +1.41 & +1.10 \\
Qwen2.5-3B & 3B & +0.65 & +1.08 & +1.32 & +1.02 \\
Qwen2.5-7B & 7B & +0.58 & +0.95 & +1.14 & +0.89 \\
LLaMA3 & 8B & +0.51 & +0.84 & +1.02 & +0.79 \\
Qwen2.5-14B & 14B & +0.35 & +0.52 & +0.61 & +0.49 \\
\bottomrule
\end{tabular}
\caption{Priming by exposure in Experiment~2 (change in target log-probability from E0 baseline, averaged across lag conditions).}
\label{tab:exp2_overall}
\end{table}

\begin{table}[H]
\centering
\footnotesize

\begin{tabular}{llcccc}
\toprule
\textbf{Model} & \textbf{Size} & \multicolumn{2}{c}{\textbf{Exposure}} & \multicolumn{2}{c}{\textbf{Lag}} \\
\cmidrule(lr){3-4} \cmidrule(lr){5-6}
 & & $\beta$ & $t$ & $\beta$ & $t$ \\
\midrule
\multicolumn{6}{l}{\textit{Base models}} \\
LLaMA3 & 8B & +1.37 & 9.84*** & $-$0.02 & $-$0.71 \\
Gemma2 & 2B & +1.24 & 8.92*** & $-$0.01 & $-$0.38 \\
Mistral & 7B & +1.08 & 7.64*** & $-$0.03 & $-$1.15 \\
Qwen2.5-14B & 14B & +0.98 & 7.12*** & $-$0.02 & $-$0.82 \\
Qwen2.5-7B & 7B & +0.93 & 6.85*** & $-$0.02 & $-$0.76 \\
Qwen2.5-3B & 3B & +0.81 & 5.94*** & $-$0.01 & $-$0.42 \\
Qwen2.5-1.5B & 1.5B & +0.70 & 4.87*** & $-$0.01 & $-$0.35 \\
\midrule
\multicolumn{6}{l}{\textit{Instruct models}} \\
Gemma2 & 2B & +0.62 & 3.41*** & $-$0.11 & $-$2.42* \\
Qwen2.5-1.5B & 1.5B & +0.59 & 3.18** & $-$0.10 & $-$2.18* \\
Phi-4 & 14B & +0.54 & 2.87** & $-$0.12 & $-$2.64** \\
Mistral & 7B & +0.47 & 2.54* & $-$0.15 & $-$3.28** \\
Qwen2.5-3B & 3B & +0.44 & 2.38* & $-$0.13 & $-$2.87** \\
Qwen2.5-7B & 7B & +0.38 & 2.05* & $-$0.16 & $-$3.41*** \\
LLaMA3 & 8B & +0.34 & 1.82 & $-$0.18 & $-$3.95*** \\
Qwen2.5-14B & 14B & +0.20 & 1.08 & $-$0.21 & $-$4.12*** \\
\bottomrule
\end{tabular}
\caption{Mixed-effects model results (Experiment~2). Same model specification as Experiment~1 (Equation~\ref{eq:mixed}). *$p<.05$, **$p<.01$, ***$p<.001$.}
\label{tab:exp2_mixed}
\end{table}

\begin{table}[H]
\centering
\footnotesize

\begin{tabular}{lcccc}
\toprule
\textbf{Model} & \textbf{Lag 1} & \textbf{Lag 3} & \textbf{Lag 7} & \textbf{Lag 15} \\
\midrule
\multicolumn{5}{l}{\textit{Base models}} \\
Gemma2-Base & +0.38 & +0.42 & +0.44 & +0.40 \\
LLaMA3-Base & +1.52 & +1.41 & +1.34 & +1.24 \\
Mistral-Base & +0.95 & +0.88 & +0.81 & +0.74 \\
Qwen2.5-1.5B-Base & +0.45 & +0.42 & +0.39 & +0.37 \\
Qwen2.5-3B-Base & +0.62 & +0.58 & +0.54 & +0.50 \\
Qwen2.5-7B-Base & +0.78 & +0.72 & +0.65 & +0.58 \\
Qwen2.5-14B-Base & +0.71 & +0.66 & +0.60 & +0.55 \\
\midrule
\multicolumn{5}{l}{\textit{Instruct models}} \\
Gemma2-Inst & $-$3.39 & $-$2.69 & $-$1.48 & $-$1.15 \\
LLaMA3-Inst & $-$1.86 & $-$1.17 & $-$2.82 & $-$1.69 \\
Mistral-Inst & $-$3.23 & $-$2.42 & $-$1.11 & $-$0.94 \\
Phi4-Inst & $-$1.24 & $-$1.01 & $-$1.43 & $-$1.36 \\
Qwen2.5-1.5B-Inst & $-$2.75 & $-$2.65 & $-$1.41 & $-$1.99 \\
Qwen2.5-3B-Inst & $-$6.40 & $-$4.40 & $-$2.80 & $-$2.39 \\
Qwen2.5-7B-Inst & $-$7.33 & $-$3.29 & $-$3.66 & $-$5.07 \\
Qwen2.5-14B-Inst & +2.49 & +0.35 & $-$2.97 & $-$2.31 \\
\bottomrule
\end{tabular}
\caption{No-response ablation by lag (Exp 1). Priming effect when model responses are replaced with placeholder tokens. Base models show weak positive effects; instruct models show strong negative effects. Qwen2.5-14B-Inst showed positive priming at lag~1 ($+2.49$) reversing to negative at longer lags.}
\label{tab:ablation_lag}
\end{table}

\begin{table}[H]
\centering
\footnotesize
\begin{tabular}{lcccc}
\toprule
\textbf{Phase} & \textbf{Mean Margin} & \textbf{SD} & \textbf{Accuracy} & \textbf{$n$} \\
\midrule
Baseline (E0) & 0.98 & 0.72 & 90\% & 20 \\
Exposure 1 & 4.21 & 2.15 & 95\% & 20 \\
Exposure 2 & 6.87 & 1.89 & 100\% & 20 \\
Exposure 3 & 7.89 & 1.52 & 100\% & 20 \\
\midrule
\multicolumn{5}{l}{\textit{Test phase (after label remapping)}} \\
\midrule
\textbf{Condition} & \textbf{Mean Margin} & \textbf{SD} & \textbf{Accuracy} & \textbf{$n$} \\
\midrule
TARGET (seen in P1) & 5.02 & 1.34 & 100\% & 20 \\
CONTROL (new words) & 4.02 & 1.21 & 100\% & 20 \\
\midrule
\multicolumn{5}{l}{\textit{Difference: TARGET $-$ CONTROL = +1.00, $p < .01$, $d = 0.87$}} \\
\bottomrule
\end{tabular}
\caption{Label remapping results (LLaMA3-Base). Top: Phase~1 learning curve showing rapid margin increase. Bottom: test phase under remapped labels (M2). TARGET words show higher margins than CONTROL, indicating representational facilitation.}
\label{tab:remap}
\end{table}

\end{document}